# Designer-User Communication for XAI: An epistemological approach to discuss XAI design

**Juliana Jansen Ferreira**

IBM Research

Rio de Janeiro, Brazil

jjansen@br.ibm.com

**Mateus Monteiro**

IBM Research

Federal Fluminense University

Rio de Janeiro, Brazil

msmonteiro@ibm.com

**Abstract**

Artificial Intelligence is becoming part of any technology we use nowadays. If the AI informs people's decisions, the explanation about AI's outcomes, results, and behavior becomes a necessary capability. However, the discussion of XAI features with various stakeholders is not a trivial task. Most of the available frameworks and methods for XAI focus on data scientists and ML developers as users. Our research is about XAI for end-users of AI systems. We argue that we need to discuss XAI early in the AI-system design process and with all stakeholders. In this work, we aimed at investigating how to operationalize the discussion about XAI scenarios and opportunities among designers and developers of AI and its end-users. We took the Signifying Message as our conceptual tool to structure and discuss XAI scenarios. We experiment with its use for the discussion of a healthcare AI-System.

**Author Keywords**

eXplainable AI; XAI design; human-AI relationship; semiotic engineering.

**CSS Concepts**

● **Human-centered computing** ● **Human-computer interaction (HCI)**● **HCI theory, concepts, and models**

**Introduction**

Explainable AI (XAI) is becoming an important feature to be considered for any AI technology. When AI is part of a high-stake decision is when XAI is necessary to enable the human-AI partnership to make a decision. The human in the loop in this human-AI partnership cannot be left out of the context to advance research about the impacts of AI on real-world problems [3][10][20]. While Machine Learning (ML) techniques and methods are resourcefully dealing with many data, humans' input adds meaning and purpose to that data [3][11]. XAI design is the bridge to provide people with an understanding of AI's outcomes, results, and even behavior to enable them to use what AI provides to make informed and conscious decisions.

End-users, with designers' and developers' collaboration, must perform a hard abstraction exercise to consider how the AI-System will be part of their practices and how it can impact and participate in their decisions. Explanations about AI should be part of this exercise's results. There is a gap between AI outputs' explanations and the explanations people need to make sense of what the AI-System did and how it can impact their actions [4][18][22]. Moreover, the societal, moral, and legal expectations of AI explanations should be discussed considering all stakeholders [7].

We are aware that explainable AI must meet the users' needs [21][22]. However, how can users be able to define and frame their own needs regarding AI explanations? Some frameworks and tools aim to enable the definition of XAI features, but a large part of them focuses on data scientists and ML developers as its end-users [2][8][16]. Some approaches, like Google PAIR [17] and IBM Team Essentials for AI framework [9], focus on supporting the discussion of XAI dimension with end-users from different domains not particularly knowledgeable about AI and its concepts. They provide guidelines to consider in XAI design, but the challenge to operationalize the discussion about XAI is still present in those approaches.

The explanation about AI's outcomes, results, and behavior has several dimensions that we should consider for XAI design. 'Who' and 'why' are two of those dimensions that have been the focus of previous research [1][15]. 'Who' are all people interested in AI explanations, like end-users, decision-makers, affected users, regulatory bodies, AI system builders. 'Why' is related to the motivations and expectations each of the interested people has for explanations. We argue that other dimensions should be considered for XAI discussion [11][12] that are not considered in the available frameworks and methods [17].

We took the Semiotic Engineering as the theoretical lens for this research [6]. We selected its conceptual tool called SigniFYIng Message [5] to structure the different dimensions that we believe should be considered for XAI scenarios' discussion. We perform an initial experiment using the SigniFYIng Message to structure XAI scenarios for a healthcare AI-System to support the discussion between the AI-System's designer and an end-user's advocate. In this experiment, we propose to add the SigniFYIng Message to operationalize the discussion of XAI between AI's designers and end-user. We present a case with IBM Essentials for AI framework, adding the SigniFYIng Message to aid the discussion in the *Knowledge activity* of the AI design process. What we learned during this experiment informed a future investigation with healthcare end-users that we could not reach at this moment due to the Covid-19 pandemic.

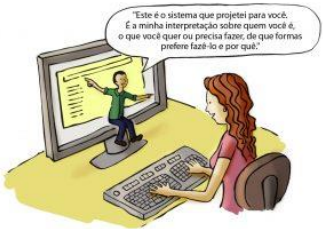

Figure 1. Metacommunication Template (adaptation from [14])

Table 1. SigniFYIng Message Description

| Dimension | Dimension Description |
|---|---|
| Who | refers to the AI designer – speaker - and the end-user in the conversation. (*Here is my understanding of who you are,…*), |
| What | refers to domain, task, interaction, interface, system, that message is about. (*…what I've learned you want or need to do, …*), |
| Why | refers to the speaker's intent relative to communication (*… in which preferred ways, and why. …*), |
| How | refers to the speaker's form of communication. (*…This is the system that I have therefore designed for you, …*), |
| When and where | refer to the speaker's context of communication. (*… and this is the way you can or should use it in order to fulfill a range of purposes that fall within this vision.*). |

**Semiotic Engineering to Frame the XAI Discussion**

We took the Semiotic Engineering [6], a comprehensive semiotic theory of Human-Computer Interaction, as the theoretical lens for this research due to its view of HCI as a particular case of computer-mediated human communication between designers and users at interaction time (Figure 1). The content of the message refers to how, when, where, and why the users can or should, themselves, communicate with the system in order to achieve certain goals and effects that are consistent with the designers/developers' vision.

With Semiotic Engineering as our theory, we are focusing on two roles each time we structure an XAI scenario for discussion: the AI-Systems designer (represents the AI-Systems development team) and the end-user for the scenario. We selected the SigniFYIng Message [5] as our epistemological tool to structure the XAI scenarios' discussion. It is a conceptual structure to frame the content of exchanged messages, not lose track of what matters and why.

The SigniFYIng Message is usually used to inspect software artifacts as part of other methods. [5] However, due to its epistemic nature, we believe it can be a valuable resource to structure the XAI scenarios considering the different dimensions and help to operationalize the discussion about XAI. The SigniFYIng Message considers the following five dimensions as described in Table 1.

**OUR CASE – MARIANA**

Our case scenario[1] involves older adults with mobility difficulties who need a multidisciplinary health team in home care service to handle chronic diseases in Brazil. It was described in a previous study [13] when health professionals co-designed a chatbot named **MarIANA**, designed to support the caregiver of older adults with hypertension. We use the provided health professionals' context and real cases as the basis for this initial experiment.

The event that starts our experiment scenario is a notification from MarIANA to the healthcare professional. During his nightshift, he receives a message from MarIANA - **"I am 80% confident that you need to contact João's caregiver. His BP is exponentially high."**. MarIANA recommended the healthcare professional who needs more information to decide if he needs to contact the patient's home immediately and if so, he must have the best orientation for the patient.

For this experiment, we built a *Reasoning statemen (*Figure 2*) for MarIANA dashboard.* This statement is the input for the last activity of IBM Essentials for AI framework – Knowledge that we considered as an interesting point to use the SigniFYIng Message to operationalize the discussion. Considering that reasoning statement, the **AI-System's designer** filled the following *SigniFYIng Message* (Figures 4-8) to discuss with the **end-user's advocate:**

---

[1] See more at this link.

> MarIANA can offer confident recommendations to aid health professionals' decision-making (About the patient health home care arrangement, i.e., medicines, equipment, caregiver plan, among others.) by showing foreseen scenarios of the patient's conditions based on **the patient medical records, patient medical records with a similar condition, MarIANA chatbot conversation records, and registers/annotations from the health professionals' visits.**

Figure 2. Experiment Reasoning statement

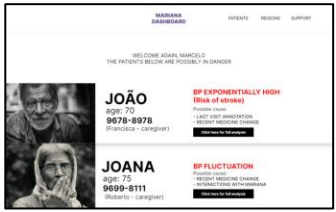

Figure 3. Prototype screen[1] generated after the discussion using the Signify Message.

> *Here is my understanding of who you are:* You are a nurse that assist on more than 20 home care patients with hypertension and works shifts at a regional hospital,

Figure 4. SigniFYIng Message - Who dimension

> *what I've learned you want or need to do*: Because of your busy routine, every time you receive a home care emergency notification by MarIANA, you need to handle to analyze the notification and sometimes other tasks at once (especially when you are on-duty) and make decisions about remote patients' health care. For that, you need to access the relevant information from a home care emergency notification quickly.

Figure 5. SigniFYIng Message - What dimension

> *in which preferred ways, and why*: Therefore, you need to understand with more detail the message from MarIANA, accessing the patient's medical history, mainly his BP measures, annotations from the last home visit, list of medication the patience is taking and a list of features that caused the notification,

Figure 6. SigniFYIng Message - Why dimension

> *This is the system that I have therefore designed for you:* For that reason, I designed a monitoring dashboard with explanations that allow you to identify and associate the data relevant for an emergency notification.

Figure 7. SigniFYIng Message - How dimension

## Final Remarks

Even without the end-users themselves involved in the discussion (the experiment involved an end-user advocate), the AI's designer perceived that he looked for a broader meaning of how the system could impact the end-user's decision-making cycle [11] in that scenario considering the whole context for the XAI

> *This is the way you can or should use it to fulfill a range of purposes that fall within this vision*: Whenever MarIANA notifies you with an emergency, that means patients are possibly in danger. I present a summary containing: a possible cause for the healthy event and the pieces of evidence that lead MarIANA to decide to notify you. You can access more detailed information: visualize the patient's blood pressure over time; records made by health professionals from that last visit; recent updates in the health care plan; and pieces of evidence from the conversations between MarIANA and the caregiver. You will also be able to provide feedback for the explanation provided.

Figure 8. SigniFYIng Message - When and Where dimension

discussion. Figure 3 illustrates part of the prototype generated after the discussion with SigniFYIng Message.

To tackle XAI Mediation Challenges [19] and bring together technical and social meanings of AI applications, we need to structure and operationalize the understanding of all roles in the AI-System. The roles that are usually represented in design discussions should have frameworks and tools that keep them aware of the "big picture" related to that AI-System that is under development and all parties affected.

What we learned during this experiment, including the filled SigniFYIng Message, will inform a future investigation with healthcare end-users, that we could not reach at this moment due to the Covid-19 pandemic. However, this experiment discussion already motivated the MarIANA's designer to reassess decisions and make changes in the previous design.

## References

[1] Alejandro Barredo Arrieta, Natalia Díaz-Rodríguez, Javier Del Ser, Adrien Bennetot, Siham Tabik, Alberto Barbado, Salvador Garcia, Sergio Gil-Lopez, Daniel Molina, Richard Benjamins, Raja Chatila, and Francisco Herrera. 2020. Explainable Artificial Intelligence (XAI): Concepts, taxonomies, opportunities and challenges toward responsible AI. Information Fusion 58: 82–115. https://doi.org/10.1016/j.inffus.2019.12.012

[2] Amazon: Amazon SageMaker Clarify Detect bias in ML models and understand model predictions. Retrieved from https://aws.amazon.com/pt/sagemaker/clarify/.

[3] Ashraf Abdul, Jo Vermeulen, Danding Wang, Brian Y. Lim, and Mohan Kankanhalli. 2018. Trends and Trajectories for Explainable, Accountable and Intelligible Systems: An HCI Research Agenda. In Proceedings of the 2018 CHI Conference on Human Factors in Computing Systems, 1–18. https://doi.org/10.1145/3173574.3174156

[4] Brent Mittelstadt, Chris Russell, and Sandra Wachter. 2019. Explaining explanations in AI. In Proceedings of the conference on fairness, accountability, and transparency, 279–288.

[5] Clarisse Sieckenius de Souza, Renato Fontoura de Gusmão Cerqueira, Luiz Marques Afonso, Rafael Rossi de Mello Brandão, and Juliana Soares Jansen Ferreira. 2016. The SigniFYI Suite. In Software Developers as Users. Springer International Publishing, Cham, 49–125. https://doi.org/10.1007/978-3-319-42831-4_3

[6] Clarisse Sieckenius De Souza. 2005. The semiotic engineering of human-computer interaction. MIT press.

[7] Finale Doshi-Velez, Mason Kortz, Ryan Budish, Chris Bavitz, Sam Gershman, David O'Brien, Kate Scott, Stuart Schieber, James Waldo, David Weinberger, Adrian Weller, and Alexandra Wood. 2019. Accountability of AI Under the Law: The Role of Explanation. arXiv:1711.01134 [cs, stat]. Retrieved February 10, 2021 from http://arxiv.org/abs/1711.01134

[8] International Business Machines Corporation: IBM Fairness 360. Retrieved from https://developer.ibm.com/technologies/artificial-intelligence/projects/ai-fairness-360/.

[9] International Business Machines Corporation: Team Essentials for AI Course. Retrieved from https://www.ibm.com/design/thinking/page/courses/AI_Essentials.

[10] Jonathan Grudin. 2009. AI and HCI: Two fields divided by a common focus. Ai Magazine 30, 4: 48–48.

[11] Juliana Jansen Ferreira and Mateus de Souza Monteiro. 2020. Do ML Experts Discuss Explainability for AI Systems? A discussion case in the industry for a domain-specific solution. arXiv:2002.12450 [cs]. Retrieved February 9, 2021 from http://arxiv.org/abs/2002.12450

[12] Juliana Jansen Ferreira and Mateus Monteiro. 2021. The human-AI relationship in decision-making: AI explanation to support people on justifying their decisions. arXiv:2102.05460 [cs]. Retrieved February 12, 2021 from http://arxiv.org/abs/2102.05460

[13] Mateus Monteiro, Luciana Salgado, Flávio Seixas, and Rosimere Santana. 2020. Co-designing Strategies to Provide Telecare Through an Intelligent Assistant for Caregivers of Elderly Individuals. In International Conference on Human-Computer Interaction, 149–166.

[14] Metacommunication Template Comic - https://sistemascolaborativos.uniriotec.br/wp-content/uploads/sites/18/2017/09/SC-cap17b.jpg (in Portuguese)

[15] Michael Hind, Dennis Wei, Murray Campbell, Noel CF Codella, Amit Dhurandhar, Aleksandra

Mojsilović, Karthikeyan Natesan Ramamurthy, and Kush R Varshney. 2019. TED: Teaching AI to explain its decisions. In Proceedings of the 2019 AAAI/ACM Conference on AI, Ethics, and Society, 123–129.

[16] Microsoft: InterpretML. Retrieved from https://github.com/interpretml.

[17] People+AI (PAIR): Explainability+Trust: Chapter Worksheep. Retrieved from https://pair.withgoogle.com/worksheet/explainability-trust.pdf.

[18] Q. Vera Liao, Daniel Gruen, and Sarah Miller. 2020. Questioning the AI: Informing Design Practices for Explainable AI User Experiences. In Proceedings of the 2020 CHI Conference on Human Factors in Computing Systems, 1–15. https://doi.org/10.1145/3313831.3376590

[19] Rafael Brandão, Joel Carbonera, Clarisse de Souza, Juliana Ferreira, Bernardo Gonçalves, and Carla Leitão. 2019. Mediation Challenges and Socio-Technical Gaps for Explainable Deep Learning Applications. *arXiv:1907.07178 [cs]*. Retrieved February 14, 2021 from http://arxiv.org/abs/1907.07178

[20] Randy Goebel, Ajay Chander, Katharina Holzinger, Freddy Lecue, Zeynep Akata, Simone Stumpf, Peter Kieseberg, and Andreas Holzinger. 2018. Explainable AI: The New 42? In Machine Learning and Knowledge Extraction, Andreas Holzinger, Peter Kieseberg, A Min Tjoa and Edgar Weippl (eds.). Springer International Publishing, Cham, 295–303. https://doi.org/10.1007/978-3-319-99740-7_21

[21] Tim Miller, Piers Howe, and Liz Sonenberg. 2017. Explainable AI: Beware of Inmates Running the Asylum Or: How I Learnt to Stop Worrying and Love the Social and Behavioural Sciences. arXiv:1712.00547 [cs]. Retrieved May 21, 2019 from http://arxiv.org/abs/1712.00547

[22] Tim Miller. 2019. Explanation in artificial intelligence: Insights from the social sciences. Artificial intelligence 267: 1–38.